\documentclass[conference]{IEEEtran}

\PassOptionsToPackage{colorlinks=true,allcolors=blue!55!black,breaklinks=true}{hyperref}
\usepackage[T1]{fontenc}
\usepackage[utf8]{inputenc}
\usepackage{microtype}
\usepackage{amsmath,amssymb,bm}
\usepackage{graphicx}
\usepackage{subcaption}
\usepackage{booktabs}
\usepackage{multirow}
\usepackage{makecell}
\usepackage{siunitx}
\usepackage{xcolor}
\usepackage{pifont}
\usepackage{enumitem}
\usepackage[ruled,vlined]{algorithm2e}
\usepackage{orcidlink}
\usepackage[numbers,sort&compress]{natbib}
\usepackage{hyperref}
\usepackage[capitalize,noabbrev]{cleveref}
\usepackage{cuted}   

\setlist{leftmargin=1.2em,itemsep=1pt,topsep=2pt}

\newcommand{\sysname}{\textsc{OrchardBench}}
\newcommand{\projurl}{\url{https://humphreymunn.github.io/orchardbench-page/}}

\newcommand{\Kp}{K_{\mathrm{p}}}
\newcommand{\Kd}{K_{\mathrm{d}}}
\newcommand{\cmark}{\textcolor{ours}{\ding{51}}}
\newcommand{\pmark}{$\sim$}
\newcommand{\xmark}{\textcolor{red!60!black}{\ding{55}}}
\definecolor{ours}{RGB}{20,110,60}

\begin{document}

\title{\sysname: A Physically-Grounded, GPU-Parallel Apple-Orchard\\
Simulation Benchmark for Agricultural Robotics}

\author{%
  \IEEEauthorblockN{Humphrey Munn\,\orcidlink{0009-0003-9771-2417}}
  \IEEEauthorblockA{%
    School of EECS, The University of Queensland\\
    \texttt{h.munn@uq.net.au}\qquad
    Project page \& code:\ \projurl}%
}

\maketitle

\begin{strip}
  \centering
  \includegraphics[width=\linewidth]{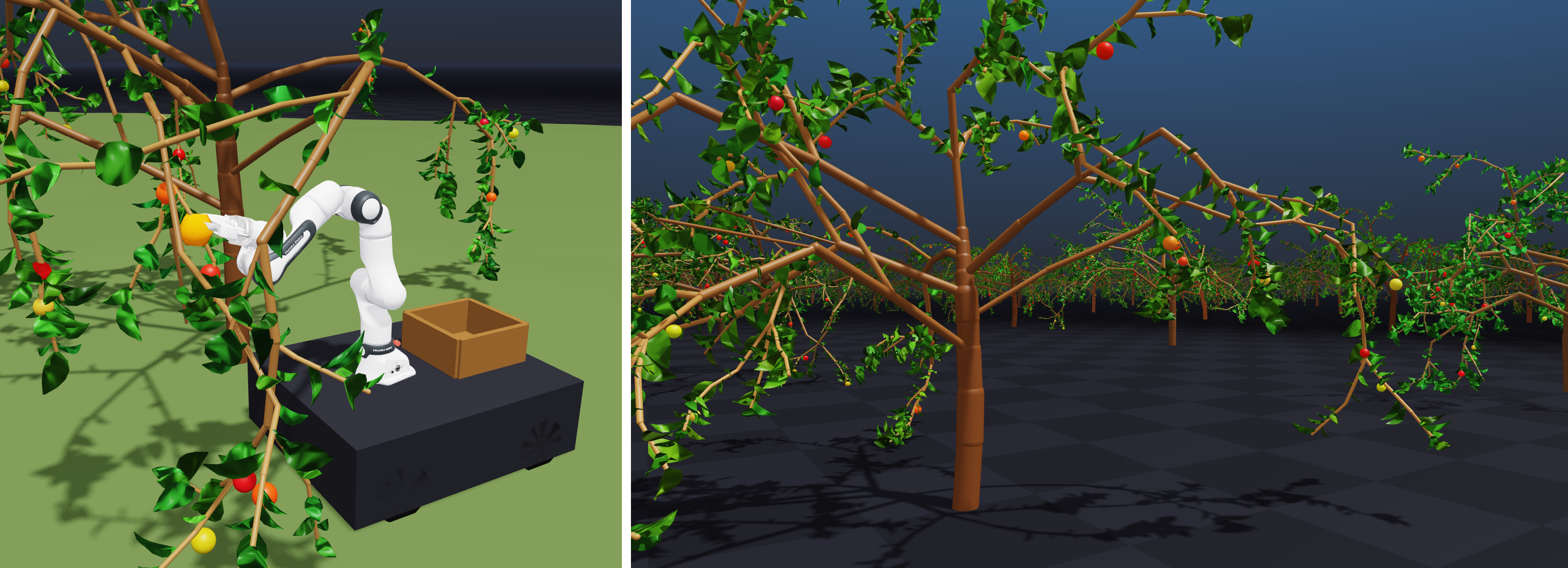}
  \captionof{figure}{\textbf{\sysname.} A physically-grounded, GPU-parallel apple
  orchard. \emph{Left:} a mobile manipulator (Clearpath Ridgeback base and Franka
  arm) autonomously harvesting a compliant, fruit-bearing tree over bumpy
  terrain---here holding a picked apple beside the collection bucket, amid foliage
  that moves with the branches. \emph{Right:} one hundred domain-randomized trees
  simulated in parallel, each a distinct plausible tree grown by a stochastic
  L-system, varying in growth habit, foliage, fruit and colour. Branches
  \emph{snap} under load and apples \emph{detach} at realistic pull forces
  (\cref{sec:physics}).}
  \label{fig:hero}
\end{strip}

\begin{abstract}
Robotic tree-fruit harvesting is a flagship problem for agricultural
automation, but progress is bottlenecked by the cost and irreproducibility of
field experiments: an orchard is available only weeks a year, every tree is
different, and a control error can permanently damage the crop or the plant.
The tree models used in graphics and agronomy are geometrically detailed but
physically inert, while the GPU-parallel simulators used in robot learning
contain no plausible trees. We present \sysname, a physically-grounded,
GPU-parallel simulation of apple-orchard trees on the \textsc{Newton}/MuJoCo-Warp
engine. Each tree is grown by a stochastic L-system and instantiated as a fully
articulated body: branches are compliant torsional spring-dampers whose
stiffness follows Euler-Bernoulli beam theory, they break at a wood modulus of
rupture and fall as free hinges, and apples are independent bodies on stem
tethers that detach at literature-grounded pull forces and load the branch as
they are pulled. A moving, density-controllable foliage layer occludes the
canopy as real leaves do. Every physical parameter is tied to a published
source. Per-environment domain randomization makes each batched world a distinct
tree, and a mobile manipulator with a wrist depth camera closes the loop with
geometric fruit perception and an autonomous harvesting baseline. Careful
engineering of the solver and the model lets \sysname\ run many parallel
environments at interactive rates on a single \SI{8}{\giga\byte} laptop GPU. We
define the tasks and a metric suite spanning harvest completeness, throughput,
and plant damage (with a per-canopy-zone breakdown), and report baseline results
across foliage, fruit load, terrain, canopy zone, and parallelism. The analytic
baseline succeeds on about \SI{40}{\percent} of the fruit it detects and harvests
only about an eighth of the reachable fruit on a tree, leaving clear headroom for
the learned and agentic methods the benchmark is built to measure. Code and videos
are released at \projurl.
\end{abstract}


\section{Introduction}
\label{sec:intro}

Tree-fruit harvesting is one of the most labour-intensive and least automated
operations in agriculture. Apples are still picked overwhelmingly by hand, the
seasonal workforce is shrinking and expensive, and the picking window for a
given block is only a few weeks long. These pressures have driven two decades of
work on robotic harvesters~\citep{bac2014harvesting,silwal2017apple,
kootstra2021selective}, but the problem remains open: fruit is heavily occluded
by foliage and by the plant's own structure, the manipulator must reach into a
compliant canopy without snapping branches or bruising fruit, and every tree is
geometrically unique. Crucially, the very thing that makes the task hard
(physical contact with a living, breakable plant) also makes it expensive to
study. A field trial requires a real orchard in season, is unrepeatable because
no two trees or approaches are identical, and carries a real cost of failure:
a mis-planned motion can tear a scaffold limb or strip fruit, damaging next
year's crop.

Simulation is the standard escape from this bind in the rest of robotics, where
GPU-parallel physics engines now train contact-rich manipulation policies at a
scale field robots could never reach~\citep{makoviychuk2021isaacgym,
todorov2012mujoco}. Orchard robotics, however, sits in a gap between two
mature but disjoint bodies of work. On one side, decades of research in computer
graphics and functional--structural plant modelling produce exquisitely
detailed trees from L-systems and related grammars~\citep{prusinkiewicz1990abop,
honda1971,costes2008mapplet}, but these models are \emph{procedural
geometry}: they describe how a tree \emph{looks}, not how it \emph{moves},
bends, or breaks under a robot's touch. On the other side, physics-based robot
learning benchmarks~\citep{james2020rlbench,yu2020metaworld,mu2021maniskill}
offer rigorous contact dynamics but populate their worlds with rigid boxes,
tools and articulated furniture, never a compliant, fruit-bearing tree. The
effort that brings the two closest together, the contact-aware
branch-manipulation work of \citet{jacob2024gentle} (PCAP), simulates a
procedural forest of compliant branches and learns to push them aside; but it
uses a spring abstraction rather than beam-theoretic dynamics and models neither
branch breakage, nor fruit, nor autonomous harvesting, and a benchmark with
standardized tasks and metrics is absent (\cref{tab:comparison}).

We close this gap with \sysname\ (\cref{fig:hero}), a simulation benchmark whose
central commitment is \emph{physical fidelity of the plant itself}. An apple
tree in \sysname\ is not a decorative mesh but a fully articulated dynamical
system: every internode is a rigid link, every branch junction a compliant
torsional spring--damper whose stiffness follows Euler--Bernoulli beam theory,
so that thin outer twigs are soft and the trunk is stiff exactly as in a real
tree. Branches \emph{rupture} when the transmitted bending moment exceeds the
wood's modulus of rupture, and the freed sub-tree falls as a genuine hinge at
gravity rate rather than vanishing or freezing. Apples are independent rigid
bodies suspended from their spurs by spring--damper stem tethers; a firm,
sustained pull, and only a firm, sustained pull, snaps the stem at a force
drawn from the harvesting literature, while a light tug merely bends the branch.
Every one of these numbers is grounded in a published source, from green apple
wood's density, elastic modulus and rupture stress to the \SIrange{14}{23}{\newton}
force needed to detach a ripe fruit (\cref{sec:physics},
\cref{tab:provenance}).

Around this plant model we build the rest of an orchard-robotics testbed. A
Ridgeback--Franka mobile manipulator carries a wrist-mounted depth camera and
executes a full autonomous harvest cycle (explore, approach, reach, grasp,
detach, deposit), driven by a geometric, learning-free fruit detector chosen for
its sim-to-real plausibility. A density-controllable \emph{foliage} layer,
rendered as leaves rigidly attached to the branch bodies so that it sways with
the tree, occludes the fruit exactly as a real canopy does, and optional bumpy
terrain adds outdoor variation. Per-environment domain randomization makes each
of the $N$ GPU-batched worlds a \emph{different} plausible tree, varying
geometry, growth habit, material, fruit load, foliage, and colour, so that a
policy or detector is never trained or evaluated on a single instance. Finally, careful
engineering of the solver and the model (a matrix-free constraint solve,
in-place rupture with no recompile, reduced-DOF fruit, and instanced foliage)
lets \sysname\ run many of these parallel worlds at interactive rates on a
single \SI{8}{\giga\byte} laptop GPU (\cref{sec:results}).

\smallskip\noindent\textbf{Why one benchmark for four communities.}\enspace
The design is driven by the observation that a physically-grounded \emph{and}
diverse \emph{and} fast orchard simulator unlocks several research programs that
each existing tool serves only partially:
\begin{enumerate}[leftmargin=1.4em,itemsep=2pt,topsep=2pt]
  \item \textbf{Safe verification of analytic controllers.} A model-based or
  solver-based picking strategy can be stress-tested against thousands of
  randomized trees, measuring exactly how often it snaps a branch or drops
  fruit, \emph{before} it is ever run on hardware, where such failures are
  costly and irreversible.
  \item \textbf{A learning and sim-to-real substrate.} The environment exposes
  the standard reinforcement- and imitation-learning interface at GPU scale,
  with domain randomization built in for transfer, so policies can be pretrained
  in simulation before fine-tuning on a real robot.
  \item \textbf{Controllable perception research.} Because foliage density,
  occlusion and lighting are dials with perfect ground truth, \sysname\ is a
  controlled testbed for fruit detection under variable foliage, a regime that
  is precisely what makes real orchard perception hard.
  \item \textbf{A real-world-grounded optimization target.} The benchmark reports
  a vector of physically meaningful metrics, throughput, success by canopy zone,
  branches snapped, fruit dropped, that a learned or automated-research method
  can be tasked to improve, with damage and safety as first-class objectives
  rather than afterthoughts.
\end{enumerate}

\smallskip\noindent\textbf{Contributions.}
\begin{enumerate}[leftmargin=1.5em,itemsep=3pt,topsep=3pt]
  \item A GPU-parallel apple-orchard simulation in which the \emph{plant} is
  physically grounded: compliant beam-theoretic branches, moment-of-rupture
  breaking with free-fall dynamics, and realistic detachable fruit that loads
  the branch, with every parameter tied to a published source.
  \item A moving, density-controllable \emph{foliage} layer that sways with the
  branches and occludes fruit as real leaves do, giving a controllable
  occlusion variable with perfect ground truth for perception research.
  \item Per-environment domain randomization of tree geometry, growth habit,
  material, foliage and appearance that keeps parallel worlds structurally
  homogeneous (and therefore GPU-batched) while making each one a distinct tree.
  \item Compute-efficiency techniques (a matrix-free constraint solve, in-place
  rupture with no model recompile, reduced-DOF fruit, instanced foliage, and an
  on-device domain-randomization build) that let many parallel environments run
  at interactive rates on an \SI{8}{\giga\byte} laptop GPU.
  \item A complete closed-loop harvesting testbed and benchmark: mobile
  manipulator, wrist depth sensing, geometric fruit perception, an autonomous
  baseline controller, and a metric suite (harvest completeness, throughput,
  plant damage, per-canopy-zone success), with baseline results across foliage,
  fruit load, terrain, canopy zone, and parallelism.
\end{enumerate}

\section{Related Work}
\label{sec:related}

\sysname\ sits at the intersection of four largely separate literatures:
procedural plant modelling, physics simulation for robot learning, agricultural
harvesting robotics, and orchard perception. Its contribution is best understood
as joining the first two, bringing physically-grounded dynamics to procedurally
generated, diverse trees, in service of the last two.

\paragraph{Procedural plant and tree modelling.}
The generation of realistic plant geometry is a mature field founded on
Lindenmayer systems~\citep{prusinkiewicz1990abop}, with the branching structure
of trees captured by Honda's parametric models~\citep{honda1971} and radii by
the pipe model of Shinozaki~\citep{shinozaki1964pipe,lehnebach2018pipe} and its
hydraulic refinements~\citep{mcculloh2003murray}. For apple specifically,
\textsc{MAppleT}~\citep{costes2008mapplet} couples a stochastic L-system to a
biomechanical model of gravimorphic bending to reproduce cultivar-specific
architecture over years of growth. These functional--structural plant models are
the state of the art in \emph{describing} how a tree is shaped, and we adopt
their machinery, a stochastic central-leader grammar with phyllotactic
divergence~\citep{okabe2015phyllotaxis} and pipe-model taper, to grow our
canopies. The essential limitation for robotics is that they are models of
\emph{static geometry}: a MAppleT tree does not bend when pushed, transmit a
moment, or break. \sysname\ takes their output as the rest configuration of a
\emph{dynamical} articulated body.

\paragraph{Physics simulation for robot learning.}
GPU-parallel physics engines have transformed robot learning by simulating
thousands of environments at once~\citep{makoviychuk2021isaacgym,mittal2023orbit},
building on accurate contact solvers such as MuJoCo~\citep{todorov2012mujoco} and
differentiable/data-parallel back-ends like NVIDIA Warp~\citep{macklin2022warp}.
A now-crowded landscape, Brax~\citep{freeman2021brax}, MuJoCo
Playground~\citep{zakka2025playground}, SAPIEN~\citep{xiang2020sapien} and
Genesis~\citep{genesis2024}, delivers ever-faster batched simulation, yet none
ships a compliant, breakable, fruit-bearing plant as an asset. We build on
\textsc{Newton}~\citep{newton2024}, a Warp-based engine whose MuJoCo-Warp solver
gives implicit, stable joint springs while running fully on the GPU. At the massive parallelism these engines are built for, however, what they are
typically used to simulate are rigid robots and rigid or articulated
\emph{objects}; compliant, high-degree-of-freedom natural structures such as
trees are rare, and a breakable, fruit-bearing one is absent. A recurring theme in our system design is making a large, stiff, sparsely
coupled articulation (a tree is $\sim$hundreds of compliant joints plus dozens of
free fruit bodies) behave stably and fast on hardware built for many small
articulations (\cref{sec:physics}).

\paragraph{Robot-learning benchmarks.}
Standardized benchmarks have been central to progress in manipulation, %
RLBench~\citep{james2020rlbench}, Meta-World~\citep{yu2020metaworld} and
ManiSkill~\citep{mu2021maniskill} among them, by fixing tasks, assets and
metrics so methods can be compared. Their scenes are deliberately generic
(blocks, tools, cabinets); none targets the specific and demanding physics of a
compliant orchard canopy, nor reports agricultural metrics such as fruit
throughput or plant damage. \sysname\ is, to our knowledge, the first benchmark
whose object of manipulation is a physically-grounded breakable plant, with a
metric suite to match.

\paragraph{Agricultural harvesting robotics.}
Robotic harvesting of tree fruit has been pursued for decades; reviews document
the recurring obstacles of occlusion, variable illumination, delicate contact and
the sheer geometric variability of plants~\citep{bac2014harvesting,
kootstra2021selective,zhang2020harvestreview}. Field systems such as the robotic
apple harvester of \citet{silwal2017apple} demonstrate a full detect--reach--%
grasp--detach cycle but also expose the cost structure that motivates us: field
integration and evaluation are slow, seasonal and unrepeatable, and controllers
that damage the tree or fruit cannot be freely iterated on live plants. A
simulator that faithfully reproduces the compliant, breakable canopy and the
detachment mechanics of the fruit lets this iteration happen safely and at scale,
which is precisely \sysname's purpose.

\paragraph{Fruit detection and perception.}
Orchard perception is a field in itself, spanning sensor surveys for fruit
detection and localization~\citep{gongal2015sensors}, the shift to deep learning
for detection and yield estimation~\citep{koirala2019deep}, and public datasets
such as MinneApple~\citep{hani2020minneapple}. The central difficulty is
occlusion by foliage and structure, which no fixed dataset can vary in a
controlled way. Because \sysname\ renders depth (and colour) with perfect
ground-truth fruit poses while foliage density is a continuous dial, it is a
controllable environment for exactly this problem, and our baseline detector is
deliberately a classical, learning-free geometric method
(\cref{sec:perception}), so that perception performance reflects the scene rather
than a pretrained network.

\paragraph{Closest work: physical tree/branch manipulation.}
The nearest precedent is the contact-aware branch-manipulation work of
\citet{jacob2024gentle} (PCAP, ``Proprioceptive Contact-Aware Policy''),
which models tree branches as rigid cylinders joined by torsional
spring--dampers with a beam-derived stiffness $\Kp = E I/\ell$ and per-level
damping, and learns a proprioceptive policy that pushes branches aside with
minimal contact force, transferring zero-shot to a real arm. We adopt their
branch-compliance formulation as our starting point and extend it substantially.
PCAP's goal is to \emph{avoid} damaging branches while reaching through them; like
us it trains on a domain-randomized procedural L-system forest, but its branch
physics is a spring abstraction and it models neither branch \emph{breakage}, nor
\emph{fruit} with detachment, nor a mobile base with fruit sensing, nor an
autonomous harvesting task with metrics. \sysname\ adds exactly these
capabilities, breaking, realistic detachable fruit that loads the whole-tree
dynamics, a mobile manipulator with depth sensing, and a benchmark task suite,
turning a branch-avoidance testbed into a harvesting benchmark. \Cref{tab:comparison} summarizes the
comparison.

\begin{table*}[t]
  \centering
  \small
  \caption{\textbf{Feature comparison} of \sysname\ against a representative
  procedural plant model, the closest physical branch-manipulation system
  (PCAP / Jacob et al.), and general-purpose GPU physics/robot-learning
  simulators. \cmark~= native/supported, \pmark~= partial/limited,
  \xmark~= absent.}
  \label{tab:comparison}
  \begin{tabular}{@{}lccccc@{}}
    \toprule
    Capability & \textbf{\sysname} (ours) & MAppleT~\citep{costes2008mapplet}
      & \makecell{PCAP / Jacob\\et al.~\citep{jacob2024gentle}}
      & \makecell{Isaac Gym/\\Lab~\citep{makoviychuk2021isaacgym}}
      & \makecell{MJX/Play-\\ground~\citep{zakka2025playground}}\\
    \midrule
    Procedural stochastic apple trees          & \cmark & \cmark & \pmark & \xmark & \xmark\\
    Pipe-model taper / botanical allometry     & \cmark & \cmark & \pmark & \xmark & \xmark\\
    Compliant articulated branch dynamics      & \cmark & \pmark & \cmark & \xmark & \xmark\\
    Branch \textbf{breaking} at rupture        & \cmark & \xmark & \xmark & \xmark & \xmark\\
    Realistic \textbf{fruit}: detach + loads tree & \cmark & \xmark & \xmark & \xmark & \xmark\\
    Per-env domain-randomized \emph{distinct} trees & \cmark & \xmark & \pmark & \pmark & \pmark\\
    GPU-parallel batched envs                  & \cmark & \xmark & \pmark & \cmark & \cmark\\
    Mobile manipulator + fruit sensing         & \cmark & \xmark & \pmark & \pmark & \pmark\\
    Autonomous harvesting task + metrics       & \cmark & \xmark & \xmark & \xmark & \xmark\\
    Lit.-grounded \emph{mechanical} parameters & \cmark & \pmark & \pmark & N/A & N/A\\
    \bottomrule
  \end{tabular}
\end{table*}

\section{Tree Generation}
\label{sec:generation}

\begin{figure}[t]
  \centering
  \begin{subfigure}{0.31\linewidth}
    \includegraphics[width=\linewidth]{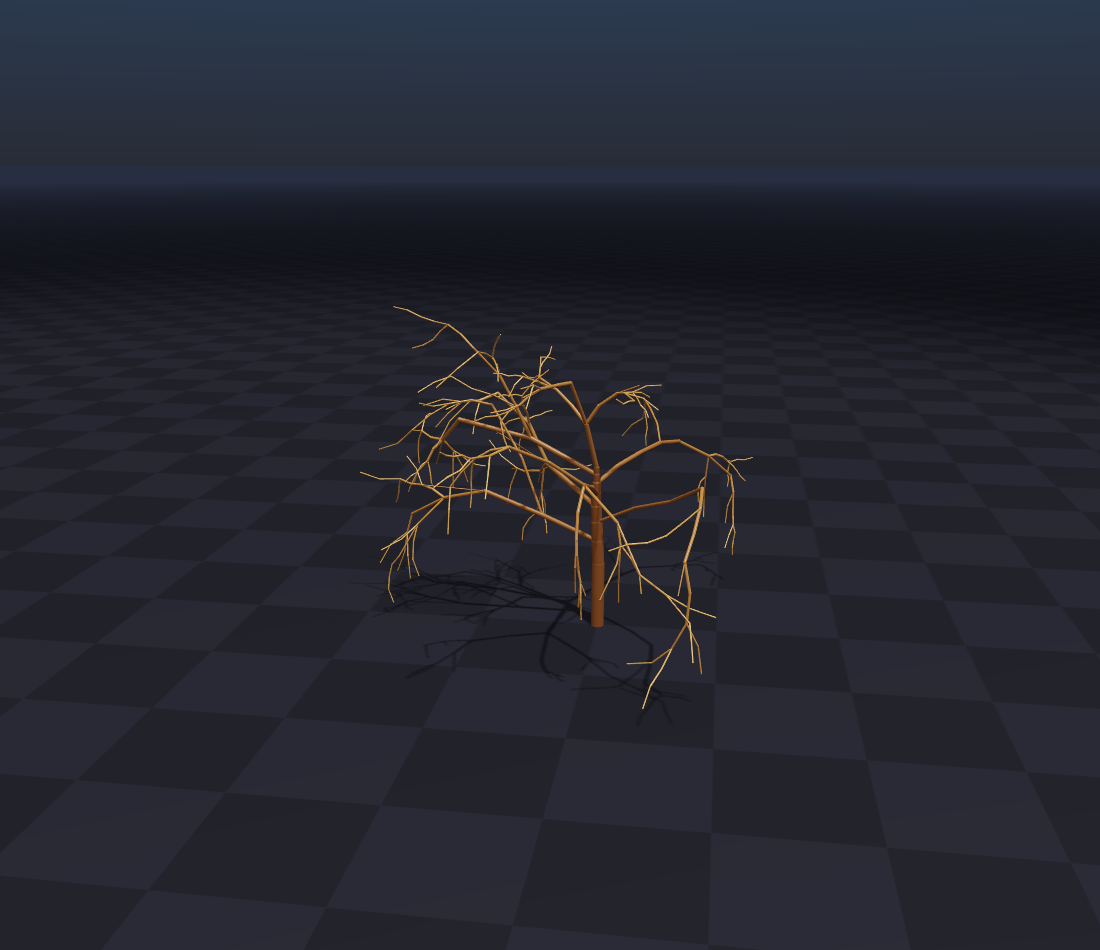}
    \caption{L-system skeleton}
  \end{subfigure}\hfill
  \begin{subfigure}{0.31\linewidth}
    \includegraphics[width=\linewidth]{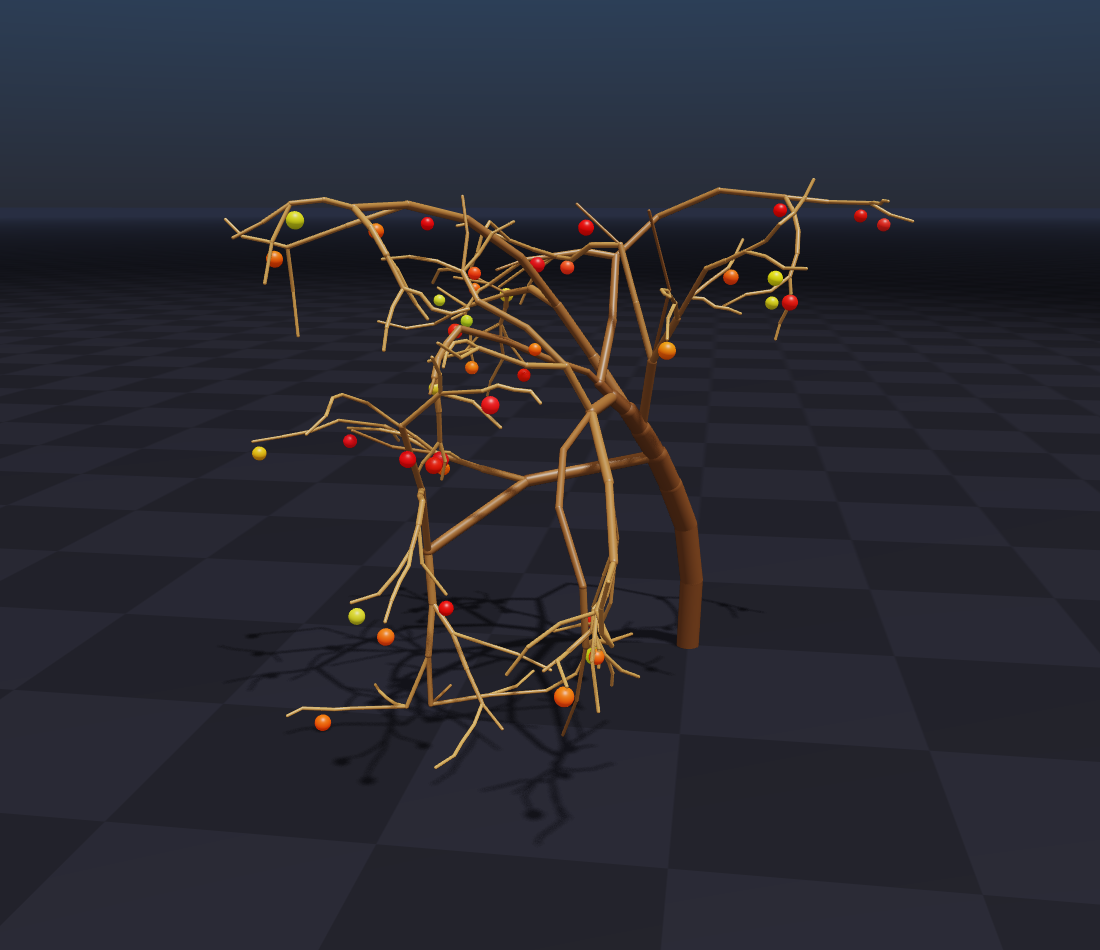}
    \caption{articulated + fruit}
  \end{subfigure}\hfill
  \begin{subfigure}{0.31\linewidth}
    \includegraphics[width=\linewidth]{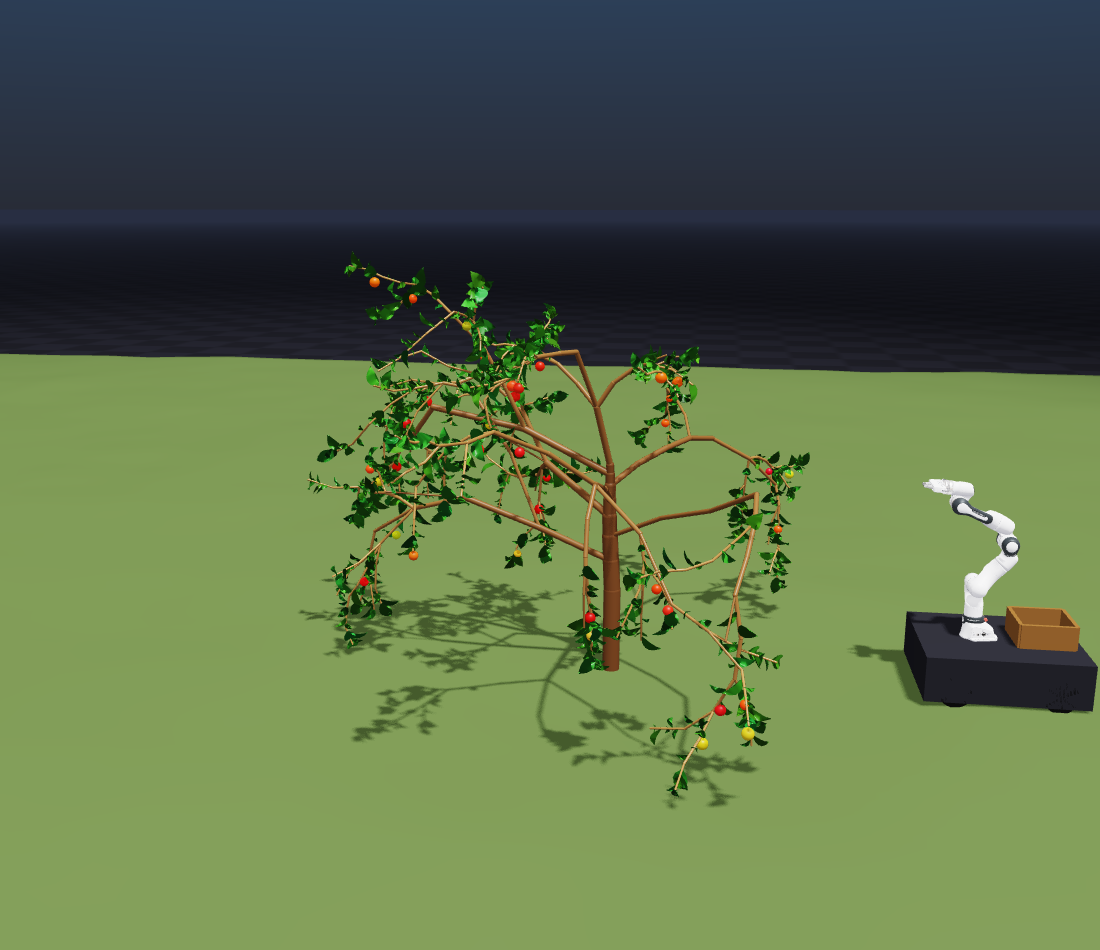}
    \caption{on terrain}
  \end{subfigure}
  \caption{From grammar to dynamical body. A stochastic L-system produces a
  skeleton of internodes (a); each becomes a rigid capsule link joined to its
  parent by a compliant joint, with foliage and fruit on the outer spurs (b);
  the finished tree, here on a heightfield terrain with the manipulator, is
  placed in the scene (c).}
  \label{fig:generation}
\end{figure}

Each tree begins as a symbolic string produced by a parametric L-system and is
interpreted by a 3-D turtle into a \emph{skeleton}: an ordered list of internode
segments, each with world-frame endpoints, a branch order, and an orientation
quaternion whose local $+z$ axis is the branch heading. The skeleton is the rest
configuration that the physics model (\cref{sec:physics}) is built to reproduce.

\paragraph{Grammar.}
We support two generators. A ternary bracketed L-system~\citep{prusinkiewicz1990abop}
with Honda-style parametric branch angles and length ratios~\citep{honda1971}
(used for reproducible, canonical trees) applies, from an axiom trunk, the
production
\begin{equation}
\begin{split}
  A \;\rightarrow\;\; & \mathtt{!}(v_r)\, F(F_0)\,
   [\,\&(a)\, F(F_0)\, A\,]\, /(d_1)\\
  & [\,\&(a)\, F(F_0)\, A\,]\, /(d_2)\,
   [\,\&(a)\, F(F_0)\, A\,],
\end{split}
  \label{eq:lsystem}
\end{equation}
where $F$ draws an internode, $\&(a)$ pitches the heading by branching angle
$a$, $/(d)$ rolls by divergence $d$, $[\,]$ push/pop the turtle state, and
$\mathtt{!}(v_r)$ scales width. Internodes elongate ($F(\ell)\!\rightarrow\!
F(\ell\, l_r)$) and thicken ($\mathtt{!}(w)\!\rightarrow\!\mathtt{!}(w\, v_r)$)
each derivation. After $n$ derivations the body count is $1+2(3^{\,n}-1)$. The
divergence angles are centred on the golden angle
$d_1\!=\!d_2\!=\!\SI{137.5}{\degree}$, the phyllotactic divergence at the shoot
apex~\citep{okabe2015phyllotaxis}.

The default generator, used throughout the benchmark, is a stochastic
apple-tree grammar in the spirit of MAppleT~\citep{costes2008mapplet}: a central
leader emits a phyllotactic spiral of \numrange{3}{6} scaffold limbs at wide
crotch angles (\SIrange{42}{65}{\degree} from vertical, the range that maximizes
limb strength and fruit-bud formation), and each limb recursively forks into a
bounded number of laterals that arch downward under simulated gravimorphism.
Every parameter, scaffold count, fork angles, per-internode droop, length
decay, bud-abortion probability, and an overall ``form'' from upright to
weeping, is drawn from a per-seed distribution, so each seed yields a distinct
but botanically plausible tree (\cref{fig:dr_gallery}).

\paragraph{Taper (pipe model).}
Radii are assigned by the da\;Vinci/Murray pipe model~\citep{shinozaki1964pipe,
mcculloh2003murray}: from a fixed tip radius $r_{\text{tip}}$, a parent radius
satisfies
\begin{equation}
  r_{\text{parent}}^{\,\beta} \;=\; \sum_{c\in\text{children}} r_c^{\,\beta},
  \label{eq:pipe}
\end{equation}
with exponent $\beta\!=\!2.2$--$2.3$. Setting $\beta\!=\!2$ is exactly
area-preserving (da\,Vinci's rule); load-bearing tree wood stays near this value
while hydraulic-optimal networks approach Murray's $\beta\!\approx\!2.5$, so
$2.0$--$2.5$ is the defensible band~\citep{lehnebach2018pipe}. The result is a
monotone taper with the trunk thickest, which, together with the beam-theoretic
stiffness of \cref{sec:physics}, produces the correct mechanical gradient from a
stiff trunk to compliant twigs.

\paragraph{Grounding and placement.}
The finished skeleton is anchored so the trunk base sits at the origin, and a
forward-kinematic ground-clearance pass lifts any branch that would otherwise dip
below \SI{\sim10}{\centi\metre}, mimicking the way real limbs grow away from the
soil. Fruit and leaves are placed on the outer canopy (\cref{sec:physics}), and
the whole tree is rescaled to a target height (\SI{2.4}{\metre} by default, a
compact, modern trained tree so that most fruit falls within the fixed-base
arm's reach envelope; \cref{tab:provenance}). Because all of these operations are
purely geometric transforms of a fixed set of segments, they preserve the graph
structure across environments, which is what makes the domain randomization of
\cref{sec:dr} compatible with GPU batching.

\section{Physical Model}
\label{sec:physics}

The defining feature of \sysname\ is that the tree is a genuine dynamical
system. Each internode segment becomes a rigid link with a capsule collider
along its local $+z$ axis; mass and inertia follow from a wood density
$\rho_w$ and the capsule geometry. The trunk base is welded to the world, and
each branch connects to its parent by a joint whose rest transform reproduces
the skeleton exactly,
\begin{equation}
  T^{\text{parent}}_{\text{joint}} = (0,0,\ell_{\text{parent}})\cdot q_{\text{rel}},
  \qquad q_{\text{rel}} = \bar q_{\text{parent}}\, q_{\text{child}},
\end{equation}
so that at rest the articulation is stress-free and matches the generated
geometry. In rigid mode the joints are fixed; in the deformable mode used
throughout the benchmark they are compliant.

\subsection{Compliant branch dynamics}
\label{sec:compliance}
Each compliant joint is a torsional spring--damper acting on two bending
degrees of freedom (a 6-DOF ``D6'' joint with the twist and translational axes
locked). Following the branch model of \citet{jacob2024gentle} and
Euler--Bernoulli beam theory, the rotational stiffness of a branch of radius
$r$ and length $\ell$ is
\begin{equation}
  \Kp \;=\; \frac{\pi}{4}\,\frac{E\, r^{4}}{\ell} \;=\; \frac{E I}{\ell},
  \qquad \Kd \;=\; c_d\,\Kp,
  \label{eq:beam}
\end{equation}
with $E$ the wood's Young's modulus, $I=\pi r^4/4$ the second moment of area of
the circular section, and $c_d$ a stiffness-proportional damping coefficient
(units of time). We stress that $c_d$ is \emph{not} a modal damping
ratio: for a joint of rotational inertia $J$ the effective ratio is
$\zeta_{\mathrm{eff}}=c_d\sqrt{\Kp}/(2\sqrt{J})$, which varies per joint with
stiffness and inertia; $c_d$ is chosen for numerical stability (so the tree
neither creeps nor rings) rather than tuned to a measured $\zeta$. Because
$\Kp\!\propto\!r^{4}$ and $r$ tapers by \cref{eq:pipe}, thin outer twigs are
orders of magnitude more compliant than the trunk: this reproduces the mechanical
behaviour of a real tree without any per-branch tuning. The
restoring torque on a joint at bending angle $\bm\theta$ (angular velocity
$\bm\omega$) is $\bm\tau=-\Kp\,\bm\theta-\Kd\,\bm\omega$. A per-level exponential
model (the ``rudimentary'' abstraction of \citet{jacob2024gentle}) is also
provided. Under \textsc{Newton}'s MuJoCo-Warp solver these are exact implicit
position actuators, which keeps the articulation (roughly $150$ internodes
$\times\,2$ bending DOF plus $\sim\!30$ fruit $\times\,3$ DOF, i.e.\
$\sim\!400$ DOF for a default apple tree) stable at the low substep counts we
use.

Wood is not perfectly elastic: we add velocity-proportional damping
$F=-c_v\, m\, v$, $\tau=-c_\omega\, I\,\omega$ (mass- and inertia-proportional so
thin twigs are not over-damped). This is a phenomenological surrogate for
aerodynamic and material losses, not literal $v^2$ air drag; together with a soft
ground penalty it lets disturbed branches settle rather than ring indefinitely.
The default $E\!\approx\!\SI{7}{\giga\pascal}$ of green apple wood is stiff enough
that hand-scale forces barely deflect a scaffold limb (physically correct), so we
report bending validation at a softer sapling modulus in \cref{sec:results}.

\subsection{Branch breaking}
\label{sec:breaking}
A branch ruptures when the transmitted bending moment exceeds the wood's
modulus of rupture $\sigma_r$ times the section modulus of a solid circular
cross-section,
\begin{equation}
  M_{\max} \;=\; \sigma_r\,\frac{\pi r^{3}}{4},
  \qquad |M| = \Kp\,\|\bm\theta\| \;>\; M_{\max}\ \Rightarrow\ \text{rupture}.
  \label{eq:break}
\end{equation}
We test the quasi-static elastic moment $\Kp\|\bm\theta\|$ (omitting the transient
damper term $\Kd\bm\omega$); an $N$-frame hysteresis then suppresses rate spikes
so a whipping branch does not chain-snap the tree. Because living branches
greenstick-fracture and buckle rather than snapping cleanly~\citep{ozden2014branches},
we place $\sigma_r$ near the low end of the green-wood range so the picker meets
realistic resistance before a branch gives.

Making a constrained body \emph{genuinely free at runtime} is the hard part on
a MuJoCo-family solver: disabling the joint welds the child rather than freeing
it, and editing the model mid-simulation forces a recompile that, under a hard
pull, injects energy and cascades into a blow-up. We avoid this entirely.
\textsc{Newton}'s MuJoCo-Warp back-end stores each joint's position-actuator
gains in device arrays of shape $(\text{worlds}, n_u)$ that are read afresh every
step; on rupture we simply zero the broken degrees of freedom's gain and bias
rows \emph{in place}, per environment, removing the spring \emph{and its
implicit damping} with no recompile and no CUDA-graph recapture. The branch
becomes a true free hinge and swings down at gravity rate. A snapped sub-tree's
linear drag is also reduced (so the surrogate damping does not fake a slow fall),
and Coulomb friction is introduced at the break and ramped up
$\sim\!5\times$ over $\sim\!\SI{1.5}{\second}$ as the torn fibres seize, so the
limb falls, swings through once, and is then pinned; it does not pendulum back
above its release height or sway forever. The entire operation is a per-joint
flag and a small array write; there is no frame-rate cost relative to a
non-breakable tree.

Applied and interactive forces are kept unconditionally stable by three rails on
the body force (impulse cap, power fade, speed brake) that never affect static
loading; we defer the details to \cref{app:stability}.

\begin{figure}[t]
  \centering
  \includegraphics[width=0.82\linewidth]{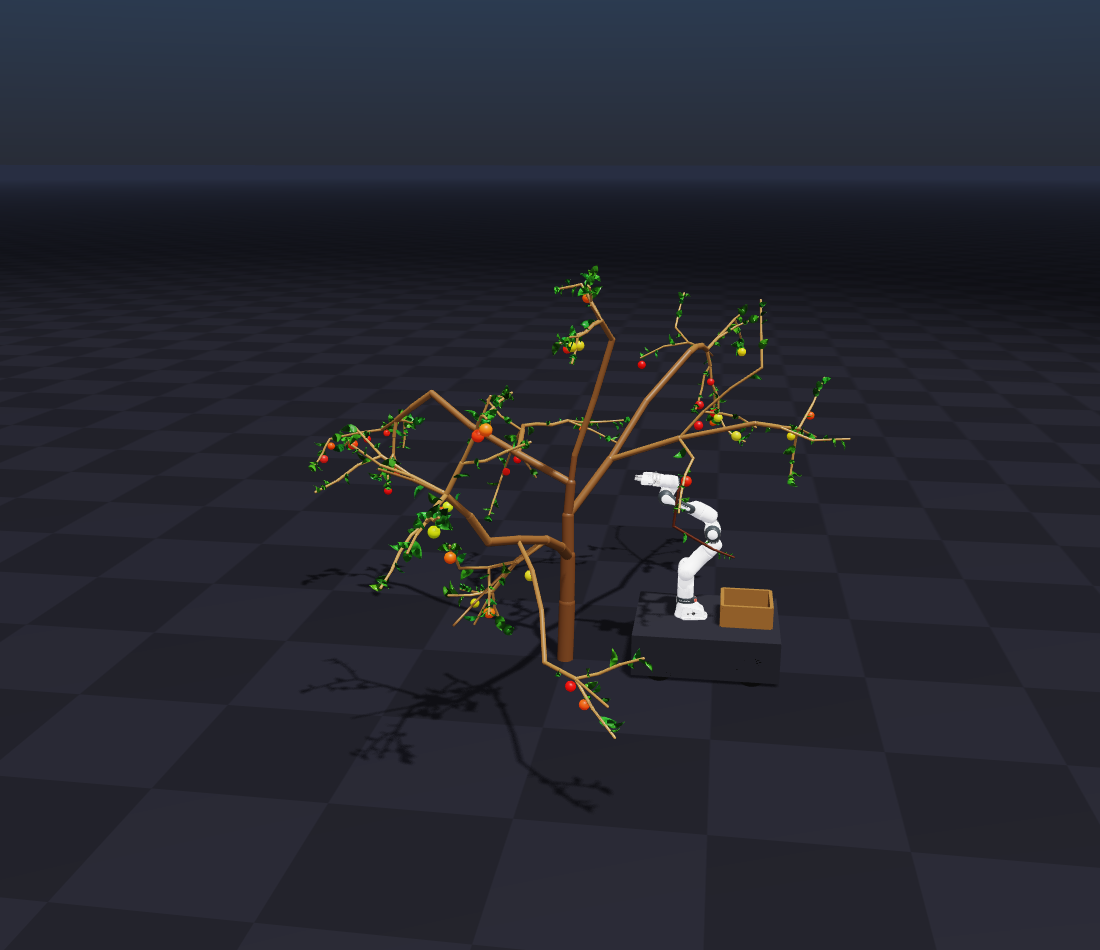}
  \caption{Branch breaking under robot contact. As the manipulator pushes into
  the canopy, a limb loaded past its modulus of rupture snaps: it goes limp and
  droops (recoloured dead-brown), and drops out of the collision set so the
  debris cannot re-excite the solver. Damage of this kind is what the
  \emph{branches-snapped} metric penalizes.}
  \label{fig:robotbreak}
\end{figure}

\subsection{Fruit}
\label{sec:fruit}
Apples are placed on eligible outer spurs with a bias toward the well-lit outer
canopy and instantiated as \emph{independent} rigid bodies, not joints of the
tree. A heavy fruit rigidly jointed to a thin compliant spur is numerically
stiff and unstable; instead each apple is held at its hang point by a one-sided
spring--damper \emph{tether}
\begin{equation}
  \bm f_{\text{tether}} = -k_s\,(\bm x - \bm x_{\text{hang}}) - c_s\,\bm v,
\end{equation}
applied to the apple only. To keep the apple's dominant cost down (each free
body is expensive in a batched solver), the default fruit body has three
translational degrees of freedom (an apple never needs to spin), which halves
its DOF count relative to a full free body.

\paragraph{Loading the branch.} A pull on the fruit must load its branch, or the
canopy would feel rigid to the manipulator. We react the \emph{elastic} part of
the tether force, minus the apple's static weight (which the tree already
carries at rest), onto the parent spur at the attach point,
\begin{equation}
  \bm f_{\text{react}} = -\big(\bm f_{s} - m g\,\hat{\bm z}\big),
  \qquad \bm f_s = -k_s(\bm x - \bm x_{\text{hang}}),
\end{equation}
clamped to a maximum so a detaching yank can never chain-snap the tree. At rest
the apple hangs a distance $mg/k_s$ below $\bm x_{\text{hang}}$, so
$\bm f_s\!=\!mg\hat{\bm z}$ and the reaction vanishes: only the elastic
\emph{perturbation} beyond rest is transmitted to the spur, and the rest pose and
rupture margins are exactly those of a tree without fruit. Under a pull the spur
visibly bends toward the hand, more on thin outer wood than on stiff scaffold.

\paragraph{Detachment.} The stem breaks along either of two paths: a direct pull
force on the fruit exceeding a threshold $f_{\text{detach}}$ sustained for a few
frames, or the stem \emph{tension} itself exceeding a slightly larger threshold
for longer; the second is what lets a gripper holding the fruit purely by
contact friction detach it, since contact forces never appear as a body force.
Stems are strong: $f_{\text{detach}}\!\in\![\,\SI{14}{},\SI{23}{}\,]\,\si{\newton}$,
grounded in measured apple detachment forces (\cref{tab:provenance}), so a light
tug only bends the branch and a whipping branch never sheds fruit. A detached
apple becomes a plain ballistic body that the drag and ground kernels settle;
detachment itself is a single flag flip with no model edit.

\subsection{Per-environment domain randomization}
\label{sec:dr}
\begin{figure*}[t]
  \centering
  \includegraphics[width=\textwidth]{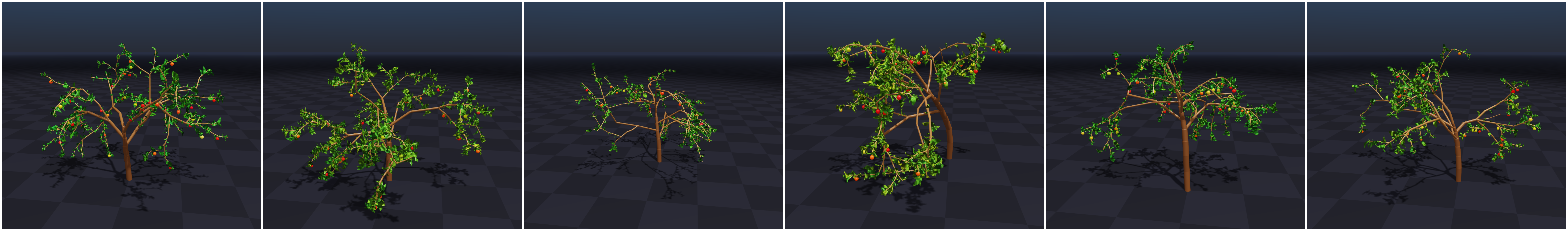}
  \caption{\textbf{Domain randomization.} A single grammar and a fixed topology,
  randomized continuously, span a diverse population of distinct but plausible
  trees varying in scale, growth habit (upright to weeping), fork spread, taper,
  foliage, fruit and coloration. Within one GPU batch the branch \emph{dimensions}
  (scale and taper) are shared across worlds so the parallel solve stays
  homogeneous and full-speed, and are resampled across batches; growth habit,
  material, fruit, foliage and coloration vary per world within the batch, and
  per-world dimensions are available as a slower option. Every world stays
  structurally identical and therefore GPU-batchable.}
  \label{fig:dr_gallery}
\end{figure*}

Domain randomization is a standard route to robust policies and sim-to-real
transfer~\citep{tobin2017domainrand,peng2018dynamicsrand}; we apply it to the
whole-tree structure. To evaluate on a \emph{population} of trees rather than one
instance, every GPU-batched environment is a different tree. The constraint is
that a batched solver requires the parallel worlds to be structurally
homogeneous (identical body, joint and shape counts and types, position by
position), while their \emph{continuous values} may differ freely. We therefore
generate one base skeleton and perturb only continuous quantities per world
(\cref{fig:dr_gallery}), holding the discrete topology fixed. One axis is treated
specially. The per-branch \emph{dimensions} (global scale and stockiness, and
per-segment length and thickness) are the only perturbation whose \emph{per-world}
variation makes the batched solve expensive: distinct dimensions across the worlds
of a step re-dimension every link and gate the parallel solve (\cref{sec:results}).
By default we therefore \emph{share} one dimension draw across the whole batch, so
the worlds stay dimensionally identical and step at the homogeneous rate, and
resample it across resets so tree shape still varies over training. Every other
axis varies per world for free, since it perturbs frames or values rather than
link dimensions: per-segment bend and whole-tree growth \emph{habit} (a
gravimorphic droop multiplier applied inside a forward-kinematic re-walk so droop
compounds per internode exactly as real arching does, plus whole-tree lean, fork
spread and an env-wide phyllotactic twist); wood density, Young's modulus and
rupture stress; fruit size, mass and visible count; leaf size and density; and
coloration (wood tint, foliage colour, and an apple-palette hue shift). Per-world
branch dimensions remain available for maximal visual variety, at a lower batched
step rate. Fruit and leaf \emph{counts} are held structurally fixed: per-tree
visible fruit count is varied by shrinking a random subset to sub-pixel size
rather than by adding or removing bodies, so the topology, and hence the batch, is
preserved. The build replicates the base environment and patches the per-world
values directly into the finalized device arrays, so startup stays fast and
cross-environment render instancing survives (in full when dimensions are shared).

We stress that fixing the topology is a \emph{within-batch} requirement for GPU
batching, not a limit on the randomization. The topology itself is a stochastic
draw of the generator: a fresh seed re-samples the scaffold count, the branching
pattern, and the whole architecture, so different runs (or re-generations)
produce genuinely different \emph{structures}, not just different continuous
values of one structure. Structural variation is therefore covered by running
several batches, each a distinct topology randomized continuously within it; a
single batch trades structural variety for the homogeneity that makes thousands
of worlds step in parallel.

\subsection{Ground, foliage, and solver}
\label{sec:extras}
\paragraph{Terrain.} An optional heightfield provides a bumpy outdoor ground:
a few octaves of value noise on a random lattice, bilinearly upsampled to a fine
grid and flattened under each trunk with a smooth disc so the tree stays planted.
For multi-env batches the noise is generated \emph{periodic} at the display-grid
pitch and tiled across the grid, and shared as a single static shape across all
worlds (batching is unaffected). Its amplitude is grounded in agricultural soil
random-roughness data (\cref{tab:provenance}); the manipulator's base rides the
surface through a terrain-following vertical servo.

\paragraph{Foliage.} Leaves are folded, curled elliptical blade meshes placed
with phyllotactic spacing on the outer twigs. Each leaf is rigidly attached to
its parent branch body, so the whole canopy \emph{moves with the tree}, swaying
as branches bend and occluding the fruit the way a real canopy does, rather than
being a static backdrop. Efficiency comes from instancing: all leaves of one of
three discrete size classes share a single mesh, so the whole canopy renders in
three draw batches regardless of leaf count, and the blades are massless and
non-colliding so their physics cost is exactly zero. Foliage density is a
continuous dial that controls leaf count and, above a threshold, populates inner
branches to occlude the tree interior: this is the controlled occlusion variable
for our perception experiments (\cref{sec:results}).

\paragraph{Solver and contact.} The MuJoCo-Warp constraint solver offers two
algorithms; we run its conjugate-gradient algorithm by default, which on a single
large articulation is $\sim\!7\times$ faster than the blocked-Cholesky algorithm
that is \textsc{Newton}'s default. On our regression suite (rest drift, static
bends, detachment, break-fall timing, and full-speed rams) the two algorithms
agree to within solver tolerance, so the speed-up does not change the physics;
\cref{sec:results} reports both step-rates. When the
manipulator is present, contact uses \textsc{Newton}'s own collision pipeline
with group filtering (tree and fruit collide with the robot but never with each
other, leaves never collide), so only a few hundred shapes reach the broad phase
and robot--tree contact is nearly free. The delicate case is the thinnest twigs: a gram-scale branch against a
\SI{130}{\kilo\gram} robot is a $\sim$$10^4{:}1$ mass ratio that an unconditioned
rigid contact cannot resolve---the usual reason such contacts are simply dropped,
which lets the manipulator pass straight through a small branch. Instead of
dropping them we \emph{condition} the ratio: each thin-twig joint carries extra
\emph{armature}---added inertia in DOF space that regularises the contact impulse
without changing the branch's static bend---and a small contact \emph{margin}
catches the contact before the fast arm has stepped through the thin capsule.
The manipulator then genuinely cannot penetrate a branch: one it cannot push
aside stalls the base, or, pushed hard enough, \emph{ruptures}, the breaking model
doubling as the physical release valve. This holds a twig centreline out of the
chassis to within measurement noise while a full autonomous pick---arm deep in
the canopy---stays stable, at a handful of extra contacts only when the arm is
actually among the branches.

\section{Manipulator, Perception and Autonomy}
\label{sec:perception}

\begin{figure}[t]
  \centering
  \includegraphics[width=\linewidth]{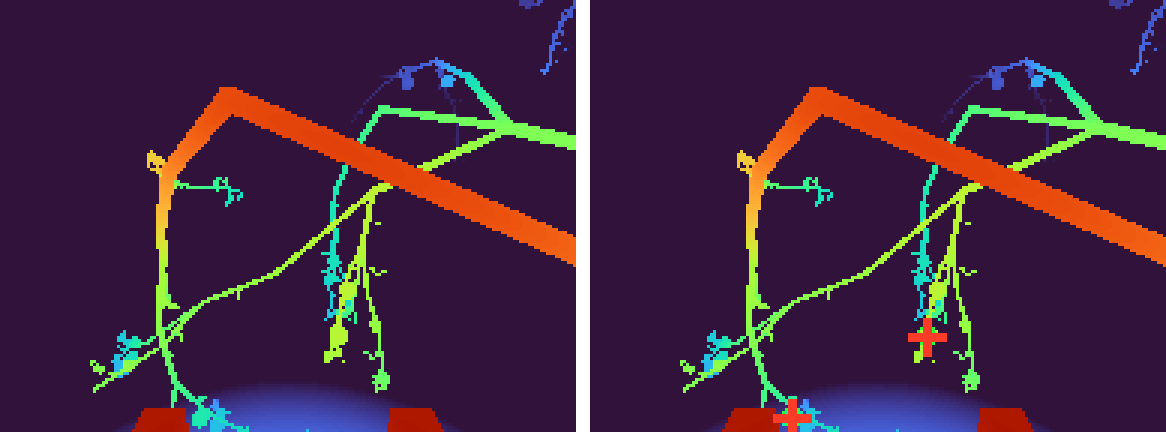}
  \caption{Wrist depth camera (left) and geometric fruit detection (right):
  depth-continuous patches are fitted with spheres and gated; accepted apples
  are back-projected to world markers, self-detections of the robot's own links
  are masked from kinematics.}
  \label{fig:perception}
\end{figure}

\subsection{Mobile manipulator}
The benchmark's embodied agent is IsaacLab's~\citep{mittal2023orbit}
RidgebackFranka: a Clearpath Ridgeback omnidirectional base carrying a Franka arm
(instantiated from the real FR3 URDF). Following common practice, the base is not simulated wheel-by-wheel
but driven through three planar degrees of freedom (world $x$, $y$, yaw) with
velocity actuators whose effort is limited to a wheel-traction-scale bound, so
the base stalls against obstacles like a real platform rather than acting as an
unbounded crusher; a fourth vertical servo tracks the terrain height. One robot
is added identically to every environment, preserving batch homogeneity. A
wrist-mounted depth camera (\textsc{Newton}'s GPU-raycast tiled-camera sensor)
renders all worlds in a single call at a realistic \SI{5}{\hertz}, with a field
of view and range matched to a consumer RGB-D sensor (\cref{tab:provenance}).

\subsection{Geometric fruit perception}
\label{sec:detector}
We deliberately use a classical, learning-free detector so that perception
performance reflects the \emph{scene}, occlusion, foliage, geometry, rather
than a pretrained network, and so that the same pipeline could run unchanged on
a real depth camera (the geometry-first localization philosophy of field
harvesters). Apples are the only sphere-like surfaces in an orchard, and a
sphere is identifiable from depth alone. The pipeline back-projects the depth
image to a point cloud, segments it into depth-continuous patches, and for each
patch fits a sphere by linear least squares: writing each point as $\bm p_i$, the
identity $\|\bm p_i-\bm c\|^2=R^2$ linearizes to
\begin{equation}
  2\,\bm p_i^\top \bm c + b = \|\bm p_i\|^2,
  \qquad b = R^2-\|\bm c\|^2,
  \label{eq:spherefit}
\end{equation}
solved for $(\bm c,b)$ and hence centre $\bm c$ and radius
$R=\sqrt{b+\|\bm c\|^2}$. Candidates are then gated on radius
(\SIrange{2}{5.8}{\centi\metre}), millimetric fit residual, convexity toward the
camera, silhouette isolation, and a \emph{pixel-count consistency} check, a
sphere of radius $R$ at range $z$ can subtend only $\approx\!\pi(Rf/z)^2$ pixels,
which rejects leaf clusters that happen to fit a small sphere. Two
\emph{anti-branch} gates reject foreshortened limb sections that fit a small
sphere alarmingly well: a footprint-elongation cap (a sphere cap is isotropic; a
branch section stretches along its axis) and an explicit sphere-versus-cylinder
residual comparison. Self-detections of the robot's own rounded links are masked
from known kinematics, and the picker adds temporal persistence before
committing. In a design-time ablation on 48 fixed canopy vantages these gates cut
wood false positives from 14 to 3 (precision $0.78\!\to\!0.95$) at a $-13\%$
single-view recall cost that the survey orbit and temporal persistence recover;
\cref{sec:results} reports precision and recall under varying foliage.

\subsection{Autonomous harvesting}
The baseline controller is an analytic state machine mirroring the pick cycle of
field systems (\cref{alg:pick}). Exploration is driven by a per-pixel
\emph{scene mask}: pixels belonging to the robot's own body or to the
ground/terrain are excluded before any ``do I see the tree?'' decision, and the
picker maintains a persistent tree-centre belief so that when the canopy leaves
view it reorients toward the remembered tree rather than rotating blind. Arm
motion is damped-least-squares inverse kinematics solved on a separate
arm-only model (a full-model solve would perturb the tree's joints), slewed
into the arm's position servos. Grasping uses a real contact grasp centred
between the fingers, whose closing force feeds the same stem-tension detachment
detector as a manual pull; the fruit is then withdrawn gently along the approach
axis until the stem gives, and deposited in a bucket on the robot's back.
Failures (unreachable target, timeout, stem too strong) blacklist the target and
return to exploration. Under multi-environment operation, an independent picker
with its own perception and metrics runs on every world, with viewer overlays on
the first; physics stays batched and per-robot host work is kept to small array
uploads.

\begin{algorithm}[t]
\small
\DontPrintSemicolon
\SetKwFor{Loop}{loop}{}{}
\KwIn{depth stream, robot state; \KwOut{fruit deposited in bucket}}
\Loop{every control step}{
  \textbf{Scan}: mask self/ground pixels; if canopy unseen, orient to belief;
  else survey-orbit for ripe fruit\;
  \textbf{Align}: drive base to a reach-aware stand-off facing the target\;
  \textbf{Reach}: DLS-IK to a pre-grasp pose behind the fruit\;
  \textbf{Grasp}: advance, close fingers, engage contact grip\;
  \textbf{Pull}: withdraw along approach axis until stem tension $>$ threshold\;
  \textbf{Deposit}: carry over bucket, release when centred and slow\;
  \lIf{unreachable / timeout / stem too strong}{blacklist, \textbf{Scan}}
}
\caption{Autonomous harvesting baseline (per environment).}
\label{alg:pick}
\end{algorithm}

\section{Benchmark Tasks and Metrics}
\label{sec:benchmark}

\sysname\ is a benchmark, not only a simulator: it fixes a set of tasks, a
metric vocabulary, and a randomized evaluation protocol so that controllers,
policies and detectors can be compared. All quantities are logged per attempt
and per environment and saved as JSON for offline analysis.

\paragraph{Tasks.} The primary task is \emph{autonomous harvesting}: starting
from a stand-off, detect, approach, grasp, detach and deposit as much fruit as
possible within a fixed episode, on a randomized tree, without breaking branches
or dropping fruit. The environment also supports component tasks that isolate
parts of the pipeline, \emph{fruit detection} (report fruit poses from the
depth stream against ground truth), \emph{reaching/grasping} a specified fruit,
and \emph{gentle interaction} (push through the canopy while minimizing
transmitted moment), and it exposes the underlying physics for open-loop
verification of analytic controllers.

\paragraph{Metrics.} An episode reports:
\begin{itemize}[leftmargin=1.3em,itemsep=1pt,topsep=2pt]
  \item \textbf{Harvest completeness (primary):} fruit deposited divided by the
  \emph{reachable} fruit present (fruit whose centre lies within the arm's reach
  sphere from an admissible stand-off). Unlike a per-attempt rate, this cannot be
  gamed by a controller that only attempts easy fruit.
  \item \textbf{Efficiency (secondary):} per-attempt place-success rate, grasp
  and detach rates, throughput (fruit deposited per simulated minute), and mean
  pick-cycle time.
  \item \textbf{Damage / loss:} \emph{branches snapped} and \emph{fruit dropped}
  (detached but not deposited), emitted unconditionally (default $0$). These are
  first-class safety costs, since a harvester that maximizes throughput by
  damaging the tree is a failure.
  \item \textbf{Effort:} maximum stem pull force applied. As a peak transient it
  can exceed the quasi-static detachment band (the grip clamp permits a brief
  overshoot); we report it as a safe-force diagnostic for a real end-effector.
  \item \textbf{Perception:} detection precision, the fraction of reported
  detections that match a real fruit (a true positive is an estimated centre
  within one apple radius of an unmatched ground-truth fruit, one-to-one
  assignment). Per-frame recall against visible ground-truth fruit is supported
  by the environment and left to future evaluation.
  \item \textbf{Compute:} simulation step rate, environment-steps per second
  available to a policy, and peak GPU memory, so results carry their cost.
\end{itemize}

\paragraph{Canopy-zone breakdown.}
Because a harvester's difficulty varies systematically over the canopy, success
is also reported by \emph{zone}. Each environment self-calibrates a zone frame
from its own reachable fruit: the trunk axis is the robust horizontal median of
fruit positions, the vertical extent is split into lower/middle/upper thirds
between the 5th and 95th height percentiles, and the radial extent is split at
the median distance from the trunk axis into inner/outer. Every pick is tagged
with its $(\text{vertical}\times\text{radial})$ zone and the reachable-fruit
census of each zone is recorded, so that per-zone success can be normalized by
the fruit actually present there (\cref{fig:zones}). This directly quantifies
the intuition that deep-inner and high fruit are hardest, which a single
aggregate success rate hides.

\paragraph{Evaluation protocol.}
Because the closed-loop task is chaotic (small perturbations in contact
ordering change which branch is nudged and whether a grasp slips), single
episodes are not meaningful. We therefore fix an official, versioned evaluation
seed set (\textsc{OrchardBench-v1}: $K$ held-out seeds, each a
domain-randomized population, disjoint from any tuning seeds) and a fixed episode
horizon (a sim-time budget; an episode ends on that budget, on all-reachable-fruit
picked, or on an unrecoverable stall). A method reports each metric as a
mean with a $95\%$ bootstrap confidence interval over trees $\times$ seeds; for
the shared-tree ablation sweeps we use paired tests and report effect sizes, and
we separate within-tree (run-to-run) from between-tree variance. This ensemble
discipline is part of the benchmark, not an afterthought. For the agentic
setting, a submission optimizes harvest completeness subject to damage
constraints ($\text{branches snapped}\le b$, $\text{fruit dropped}\le d$) and
reports the achieved Pareto point; the leaderboard ranks on constrained
completeness. A one-line mapping of these metrics onto the four use cases of
\cref{sec:intro} is deferred to that section.

\section{Experiments}
\label{sec:results}

\paragraph{Setup.} Unless noted, experiments use the stochastic apple generator
with breaking enabled, the autonomous baseline of \cref{sec:perception}, and
$N$ domain-randomized environments run in parallel on a single NVIDIA RTX~2000
Ada laptop GPU (\SI{8}{\giga\byte}). Each sweep fixes the random seed so the
\emph{same} population of trees is reused across conditions, making every sweep a
paired comparison in which only the swept variable changes. Metrics are pooled
across the parallel trees and reported with $95\%$ bootstrap confidence
intervals; the canopy-zone table pools across all twelve autonomous runs
(80 trees). \Cref{tab:headline} gives the nominal-condition baseline. Each robot
deposits about \num{1.9} fruit per minute, below the $60/7.3\approx8$ per minute
that a continuous \SI{7.3}{\second} pick cycle would allow, because much of the
episode is spent searching, on failed attempts, and in recovery; throughput is
reported per robot (the parallel environments are independent single-robot
trials, not one multi-robot stand).

\begin{table}[t]
  \centering\small
  \caption{Baseline autonomous-harvesting results (nominal condition:
  \num{40} apples, foliage density $0.6$, no terrain; pooled over 18
  domain-randomized trees, 3 seeds). The analytic baseline leaves substantial
  headroom, especially in fruit dropped, which is the point of a benchmark.}
  \label{tab:headline}
  \begin{tabular}{@{}lc@{}}
    \toprule
    Metric & Value \\
    \midrule
    Success per detected fruit (per attempt) & $0.41$ \\
    Harvest completeness (picked / reachable) & $0.12$ \\
    Throughput per robot (fruit/min)        & $1.9$ \\
    Mean pick-cycle time (s)                & $7.3$ \\
    Detection precision                     & $0.90$ \\
    Branches snapped per tree               & $\sim\!0.2$ \\
    Fruit dropped per tree                  & $\sim\!6$ \\
    Max stem pull force (\si{\newton}, peak) & $29$ \\
    \bottomrule
  \end{tabular}
\end{table}

\subsection{Parallelism and cost}
\begin{figure}[t]
  \centering
  \includegraphics[width=\linewidth]{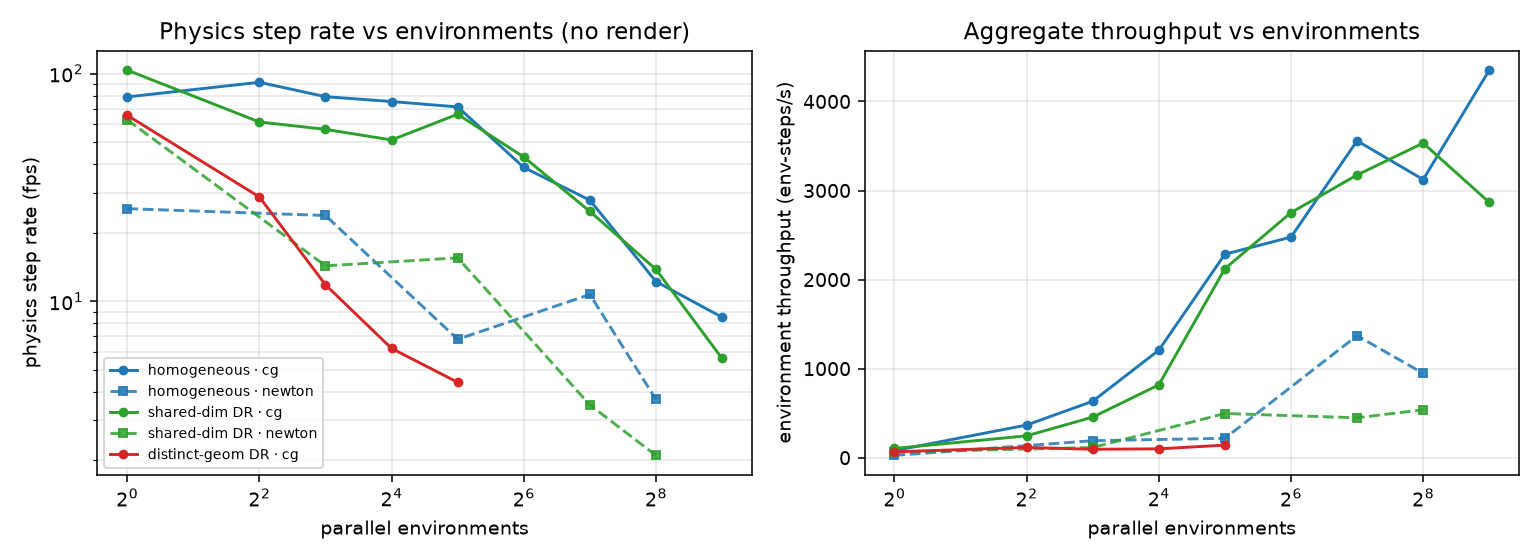}
  \caption{Physics step rate (left) and aggregate environment throughput (right)
  versus the number of parallel trees (no rendering), for three randomization
  modes and both constraint solvers. \emph{Shared-dimension} DR (green) tracks the
  \emph{homogeneous} batch (blue) and both scale to \num{512} trees, whereas
  \emph{distinct-geometry} DR (red) flatlines as its heterogeneous per-world
  geometry breaks the batched solve; CG (solid) far outscales the Newton solver
  (dashed). Single laptop GPU; each step-rate cell is a median over repeats to
  average run-to-run clock variation.}
  \label{fig:scaling}
\end{figure}

For learning research the key question is how many environments the benchmark can
run at once. \Cref{fig:scaling} sweeps parallel trees physics-only (no rendering) across
three randomization modes and both constraint solvers. A single tree steps at
\SIrange{60}{100}{} fps (CG, depending on the drawn shape); as environments are
added the per-step rate falls but aggregate throughput keeps climbing.
Homogeneous (non-randomized) batches and \emph{shared-dimension} domain-randomized
batches (our default, \cref{sec:dr}) track each other closely: both reach
\num{512} trees and plateau near \SIrange{3500}{4400}{} environment-steps/s,
confirming that shared-dimension DR steps at essentially the homogeneous rate.
\emph{Distinct-geometry} DR (per-world branch dimensions) does not scale---its
throughput flatlines near \SI{140}{} environment-steps/s and its step rate
collapses (\SI{66}{} versus \SI{4.4}{} fps at \num{32} trees, a $\sim\!15\times$
gap)---because heterogeneous kinematic geometry across worlds breaks the batched
solve. The conjugate-gradient solver reaches \num{512} trees
($\sim$\num{170000} bodies) using only $\sim$\SI{1.5}{\giga\byte} of the
\SI{8}{\giga\byte} laptop GPU, so the throughput plateau is compute-bound with
ample memory headroom; the default blocked-Cholesky (``Newton'') algorithm caps
at \num{256} trees, runs $2$--$3\times$ slower and uses roughly twice the memory,
so the matrix-free CG solve is what makes laptop-scale batches practical. In-place
rupture and reduced-DOF fruit keep breaking and fruit essentially free; the
bottleneck at scale is the constraint solver, not asset storage.

Two mechanisms underlie this, one at build time and one per step. The build:
computing each world's distinct geometry with a host-side forward-kinematic
re-walk is $O(N)$, capping distinct-geometry builds at a few tens of trees. We remove this ceiling with an \emph{on-device} build
that perturbs every world's branch geometry in a single Warp kernel (one thread
per world; fruit and foliage re-homing remain on the host path); it reproduces
the replicated base model exactly at zero strength and
produces stable, diverse trees at a build cost comparable to a homogeneous batch
(e.g.\ \SI{6.5}{\second} versus \SI{29}{\second} for the host path at \num{64}
trees, and \SI{22}{\second} at \num{512} where the host path takes minutes).
Second, and independent of the build path, is a \emph{per-step} cost, but we find
it is confined to a single randomization axis. Decomposing the per-world
perturbation and measuring each in isolation (\num{8} settled worlds), randomizing
per-world \emph{mass}, joint \emph{stiffness} and \emph{damping} runs at the
homogeneous step rate over wide ranges, individually and combined
(\SIrange{100}{133}{} fps against $\sim$\SI{110}{} fps for identical worlds), as
does randomizing the growth-habit \emph{angles} (droop, lean, spread, twist;
$\sim$\SI{83}{} fps) and all \emph{fruit and foliage} attributes (apple size,
mass, count and colour; leaf size and colour), since fruit are independent free
bodies and leaves are massless render-only cards. The only expensive axis is
per-branch \emph{size}, segment length, thickness, and global scale, which
re-dimensions every link and drops the batch to \SIrange{8}{13}{} fps; the slowdown persists after the trees settle to
rest, so it is a solver cost of the heterogeneous \emph{kinematic geometry}, not a
transient. The practical consequence is favourable: the standard reinforcement-%
learning randomizations (dynamics and gains) and the visually dominant shape
variation (growth habit) are essentially free at scale, and only re-dimensioning
each branch gates the batched solve. Crucially, this cost is incurred only when
the per-branch size differs \emph{simultaneously} across the worlds of a single
batched step; it is not a cost of varying size at all. Size variation is
therefore recovered at scale by holding one size draw \emph{shared} across the
batch, so every step stays dimensionally homogeneous and runs at the full rate,
and resampling that draw \emph{across resets} so the training distribution still
spans diverse shapes over time. A large diverse batch can thus randomize
dynamics, materials, habit and posture per world at full speed while sharing one
branch-size realization across the batch, trading within-batch geometric
decorrelation, at the cost of a per-reset rebuild, for a homogeneous, full-speed
step. The rates above are physics-only: unlike the solve, rendering is not
batched across worlds, so visualising a large batch on-screen is far slower (a
windowed view of \num{100} worlds drops to single-digit fps). Large-batch data
collection therefore runs headless, with rendering reserved for inspecting a
handful of worlds at a time.

\subsection{Effect of foliage (perception under occlusion)}
\begin{figure}[t]
  \centering
  \includegraphics[width=\linewidth]{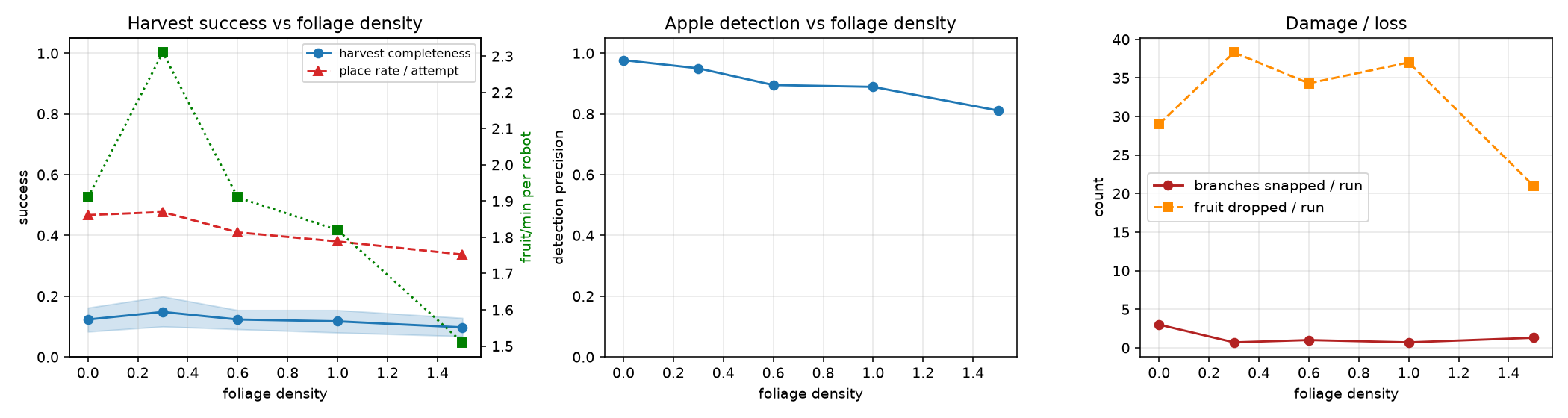}
  \caption{Foliage sweep: harvest success and per-robot throughput (left),
  detection precision (centre), and damage/loss (right) versus foliage density.
  Both detection precision and picking success degrade as foliage occludes the
  fruit. Shaded band: $95\%$ bootstrap confidence interval over trees
  (3 seeds $\times$ 6 trees per point).}
  \label{fig:foliage}
\end{figure}

Foliage density is swept with all else fixed, so the same trees are seen through
increasing occlusion. As density rises from $0$ to $1.5$, detection precision
falls from $0.98$ to $0.81$ (more leaf clutter produces more false positives
that the geometric gates must reject) and the per-attempt place rate falls from
$0.47$ to $0.34$ (\cref{fig:foliage}). This is the variable-foliage perception
and manipulation difficulty the moving foliage layer is designed to expose, here
with perfect ground truth.

\subsection{Effect of fruit load and terrain}
\begin{figure}[t]
  \centering
  \begin{subfigure}{\linewidth}
    \includegraphics[width=\linewidth]{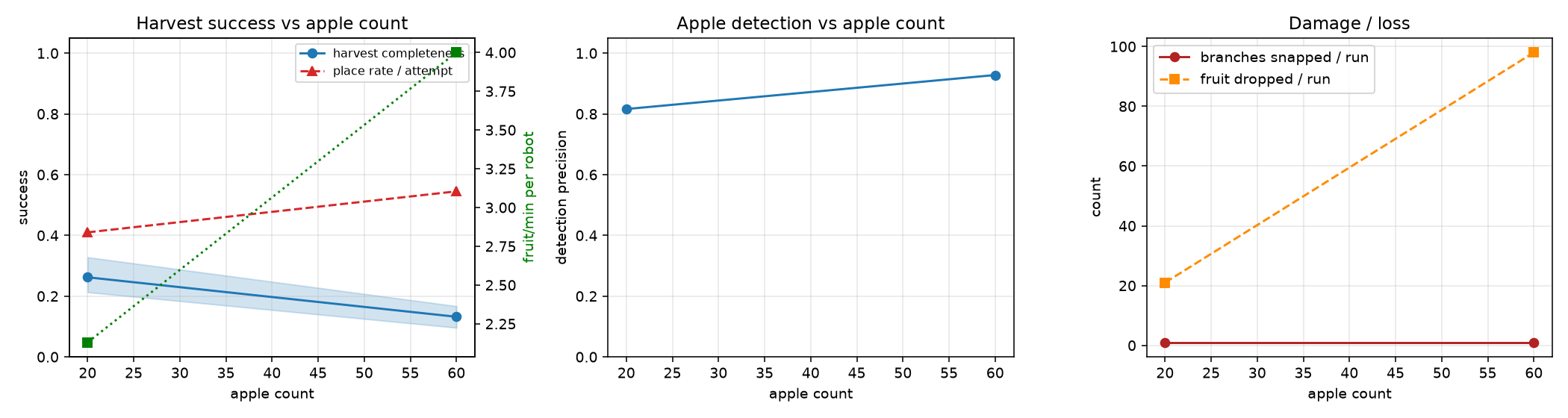}
    \caption{Fruit-load sweep (20--60 apples).}
    \label{fig:apples}
  \end{subfigure}\\[2pt]
  \begin{subfigure}{\linewidth}
    \includegraphics[width=\linewidth]{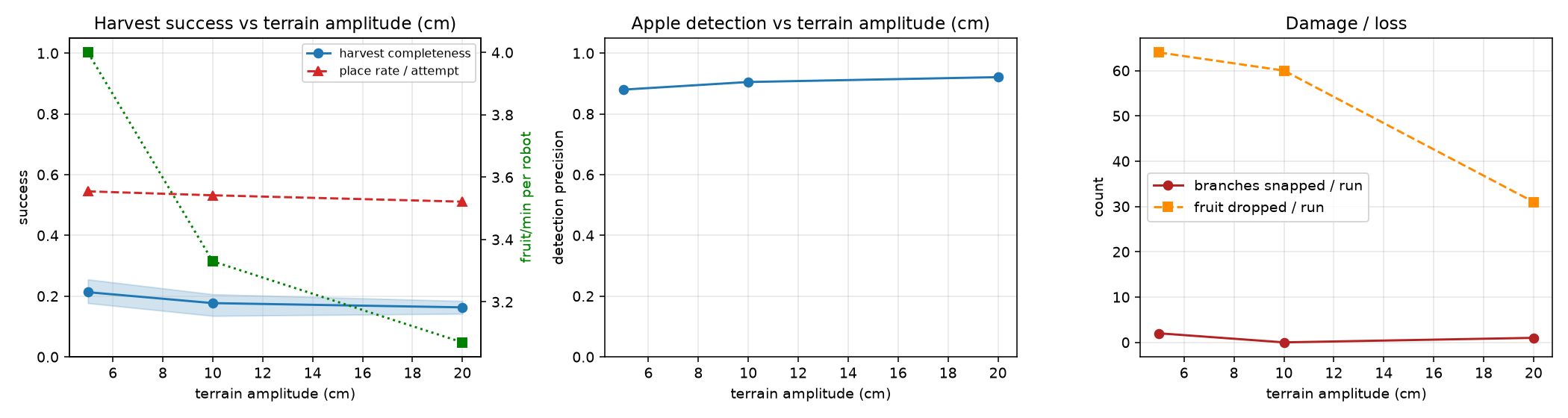}
    \caption{Terrain-amplitude sweep (5--20\,cm).}
    \label{fig:terrain}
  \end{subfigure}
  \caption{Sweeps over fruit load and terrain roughness.}
\end{figure}

\begin{figure}[t]
  \centering
  \includegraphics[width=0.80\linewidth]{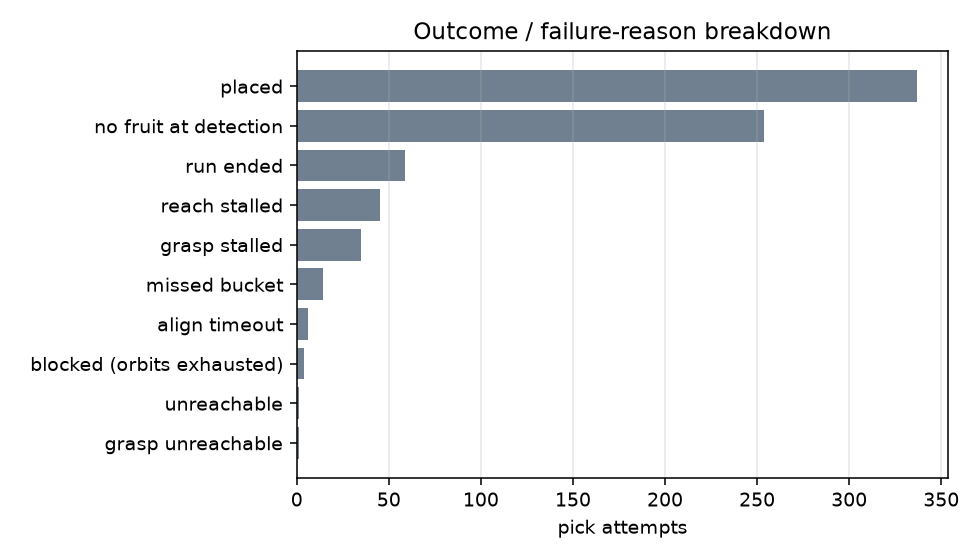}
  \caption{Pooled outcome and failure-reason breakdown. Most misses are
  false-positive detections and grasp stalls rather than navigation, which makes
  each sweep interpretable.}
  \label{fig:failures}
\end{figure}

\Cref{fig:apples,fig:terrain} sweep the number of apples and the terrain
amplitude. More fruit raises per-robot throughput (\SIrange{2.1}{4.0}{} fruit/min
from \num{20} to \num{60} apples) but lowers completeness ($0.26$ to $0.13$),
since a denser tree cannot be cleared within the episode horizon. Terrain up to the
\SI{20}{\centi\metre} orchard-alley ceiling remains drivable: the per-attempt
place rate falls only mildly, from $0.55$ at \SI{5}{\centi\metre} to $0.51$ at
\SI{20}{\centi\metre}, so navigation on bumpy ground is not the dominant failure
mode. The failure-reason breakdown (\cref{fig:failures}) attributes most misses
to false-positive detections (\emph{no fruit at detection}) and grasp stalls
rather than navigation.

\subsection{Success by canopy zone}
\begin{figure}[t]
  \centering
  \begin{subfigure}{0.46\linewidth}
    \includegraphics[width=\linewidth]{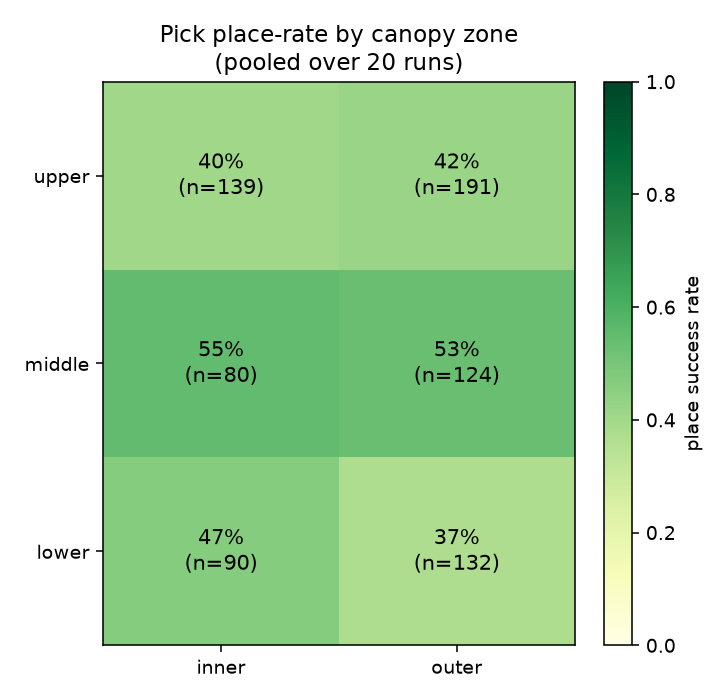}
    \caption{place-rate by zone}
    \label{fig:zones}
  \end{subfigure}\hfill
  \begin{subfigure}{0.46\linewidth}
    \includegraphics[width=\linewidth]{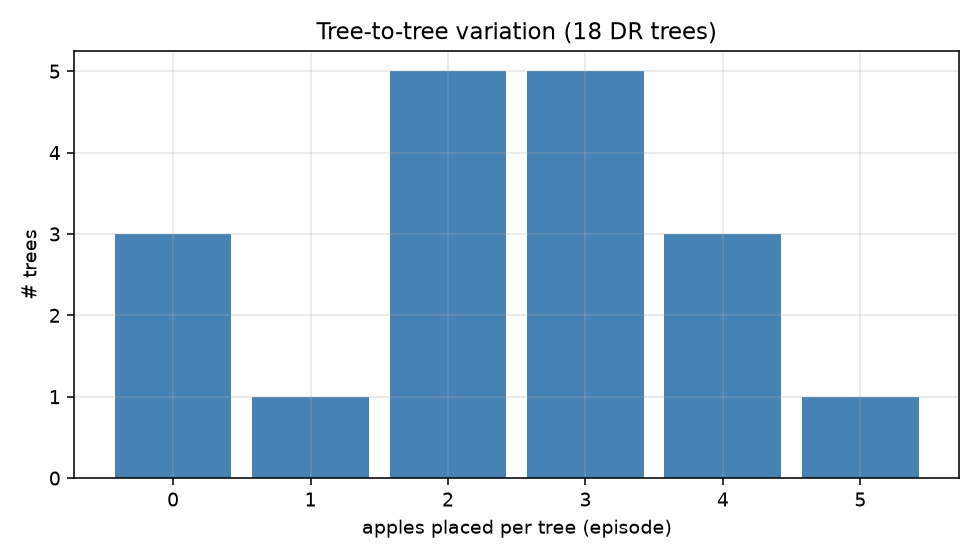}
    \caption{per-tree spread}
    \label{fig:variation}
  \end{subfigure}
  \caption{(a) Place-success rate over the $3\times2$ canopy grid, colour-coded,
  with per-zone attempt counts. (b) Histogram of apples deposited per tree across
  the 18-tree population: most trees yield one to four, illustrating the
  tree-to-tree and run-to-run spread that motivates population-level evaluation
  with confidence intervals.}
\end{figure}

\Cref{fig:zones} reports place-success over the $3\times2$ canopy grid, pooled
across all runs (80--191 attempts per zone). Success is highest in the middle
canopy ($\sim\!0.55$) and lowest in the lower-outer and upper zones
($0.37$--$0.42$), matching the intuition that fruit hanging below the outer
canopy and high on the tree are hardest to reach and see.

\subsection{Tree-to-tree variation and repeatability}
Across the 18 trees of the baseline and extra-seed runs, the number of apples
deposited per tree is $2.4\pm1.5$ (mean\,$\pm$\,s.d., range $0$--$5$;
\cref{fig:variation}). This spread, together with the chaotic per-episode
dynamics, is why we evaluate on a population and report confidence intervals: a
single episode on a single tree would not be representative.

\section{Discussion}
\label{sec:discussion}

The baseline results sketch the shape of the problem \sysname\ poses. Perception
is the dominant bottleneck: the most common failure is a committed detection with
no fruit actually there, and detection precision falls steadily as foliage
occludes the canopy, dragging picking success down with it. The analytic
baseline also exposes a sharp damage-versus-throughput tension: it succeeds on
about two in five of its attempts but knocks roughly six fruit per tree to the
ground in the process, so a method that merely maximized throughput would score
poorly on the damage metrics that a real grower cares about. The canopy-zone profile localizes
the difficulty to the deep-inner and upper canopy, suggesting that better
viewpoint planning and reach strategies, not faster motion, are where the gains
are. These are exactly the levers a learned or agentic method would tune, and the
metric suite is designed to reward doing so without trading away plant safety.

\subsection{Limitations}
\label{sec:limitations}
We are deliberately conservative about what \sysname\ does and does not
establish. First, the physics is \emph{grounded} in the literature, not
\emph{validated} against physical branch-break or grasp experiments: the
Euler--Bernoulli torsional-spring lumping is an idealisation, and the green-wood
modulus and rupture stress are reduced estimates from dried-wood references
rather than direct measurements on live apple branches (\cref{tab:provenance}).
The detachment force is set within, not calibrated to, a broad
(\SIrange{9}{40}{\newton}) reported spread. Establishing quantitative agreement
with real measurements is important future work, and until then results should be
read as \emph{relative} comparisons within the simulator rather than absolute
predictions of field performance. Second, the fruit model omits pedicel
\emph{twisting}: detachment is triggered by pull and stem tension but not by the
twist-and-pull motion many pickers use. Third, sensing is depth-only and the
renderer is not a validated photometric model, so RGB-based perception transfer
is out of scope here. Fourth, and most importantly, \emph{sim-to-real transfer is a design goal
we support, not a result we demonstrate}: no real-robot experiment is reported,
and the domain randomization is motivated by, not shown to achieve, transfer.
Finally, we position \sysname\ as a substrate for learned policies and agentic
optimization and report the environment throughput available to a policy, but we
evaluate only the analytic baseline here; a learned or agentic baseline is future
work, and the four use cases should be read as the interface we provide rather
than as validated results.

\subsection{Future work}
\label{sec:future}
\sysname\ is designed as a substrate that these extensions slot into without
re-architecting:
\begin{itemize}
  \item \textbf{Learned policies and perception.} The GPU-batched, randomized
  environment is ready for reinforcement- and imitation-learning of harvesting
  policies and for learned (RGB-D) detectors, with our analytic controller and
  geometric detector as baselines to beat.
  \item \textbf{Sim-to-real validation.} The most valuable next step is closing
  the loop with a real Ridgeback--Franka in an orchard or on cut branches, to
  calibrate the wood and detachment parameters and measure the transfer gap:
  this is the experiment that would upgrade our claims from grounded to validated.
  \item \textbf{Fruit twisting.} Adding a twist degree of freedom and a
  torsion-dependent detachment criterion to the stem model is a small extension
  that would capture the twist-and-pull strategy of human and robotic pickers.
  \item \textbf{Photorealistic rendering.} A validated photometric renderer
  (realistic foliage and lighting) would extend the benchmark to RGB perception
  and appearance-based sim-to-real, if the target application requires it.
  \item \textbf{Other crops and structures.} The generation and physics pipeline
  is not apple-specific; trellised vines, stone fruit, and other trained systems
  are natural targets, making \sysname\ a template for a family of
  physically-grounded agricultural benchmarks.
  \item \textbf{A benchmark for automated research.} Because the task is drawn
  from a real agricultural problem and the metric vector explicitly includes
  damage and safety, \sysname\ is a real-world-grounded target for automated
  research systems that co-optimize perception and control to push throughput up
  while holding branch breakage and fruit drop down.
\end{itemize}

\subsection{Conclusion}
\label{sec:conclusion}
\sysname\ brings the physical realism of a compliant, breakable, fruit-bearing
tree, long the missing ingredient, into a GPU-parallel, domain-randomized
simulation that runs on a laptop, and packages it as a benchmark with a metric
suite matched to what agricultural harvesting actually requires: not just
success, but throughput at bounded damage. Our analytic baseline clears only a
fraction of the reachable fruit and drops more than it keeps, leaving ample room
for the learning, perception, and agentic methods the benchmark is built to
measure. By grounding every physical parameter in the literature and releasing
the environment openly, we aim to let these communities iterate on orchard
robotics safely and at a scale the field itself can never offer.

\bibliographystyle{unsrtnat}
\bibliography{orchard}

\appendices
\section{Parameter Provenance}
\label{app:provenance}

A central design principle of \sysname\ is that physical parameters are not
free knobs but are tied to published measurements, so that the simulator's
behaviour is defensible and its numbers are meaningful across the four target
use cases. \Cref{tab:provenance} lists the principal parameters, their default
values, and their source. Where a value is a deliberate simulation compromise
(e.g.\ tree height or apple size, chosen for the arm's reach envelope or the
gripper's opening), the note states so and gives the real figure; where a
value is a numerical/solver setting rather than a physical one, it is marked as
such. Green-wood elastic modulus and rupture stress are reduced estimates from
dried-wood references (green wood runs $\sim$\SIrange{10}{40}{\percent} below the
12\%-moisture values) and should be treated as literature-grounded ranges rather
than validated constants (\cref{sec:limitations}).

\begin{table*}[t]
  \centering\small
  \caption{Principal physical parameters, defaults, and their provenance. Ranges
  in brackets are the domain-randomization or literature spread. ``sim.\ choice''
  marks deliberate compromises; ``solver'' marks numerical-stability settings.
  Entries with a citation are grounded in that source; parenthetical notes marked
  ``nominal'' are representative values from standard horticultural practice or
  manufacturer datasheets, chosen as reasonable defaults rather than measured
  constants.}
  \label{tab:provenance}
  \begin{tabular}{@{}p{0.26\textwidth} l p{0.46\textwidth}@{}}
    \toprule
    Parameter & Default (range) & Source / justification\\
    \midrule
    \multicolumn{3}{@{}l}{\emph{Wood material (green apple, \emph{Malus domestica})}}\\
    Density $\rho_w$          & \SI{850}{\kilo\gram\per\metre\cubed} ([780,1000]) & Basic SG 0.61; \cite{fpl2010woodhandbook,wooddatabase_apple}\\
    Young's modulus $E$       & \SI{7}{\giga\pascal} ([6,8]) & Dried $\approx$\SI{8.8}{\giga\pascal}; green $\sim$10--25\% lower \cite{fpl2010woodhandbook,niklas2011green,kretschmann2010greenprops}\\
    Modulus of rupture $\sigma_r$ & \SI{50}{\mega\pascal} ([45,60]) & Dried $\approx$\SI{88}{\mega\pascal}; green $\sim$35--40\% lower; low end for greenstick tolerance \cite{fpl2010woodhandbook,ozden2014branches}\\
    Stiffness-prop.\ damping coeff.\ $c_d$ & \SI{0.1}{\second} & \emph{Solver} value ($\Kd\!=\!c_d\Kp$; \emph{not} a modal ratio, see \cref{eq:beam}); physical branch $\zeta\!\approx\!0.01$--0.075 \cite{james2014branchdynamics}\\
    \midrule
    \multicolumn{3}{@{}l}{\emph{Morphology}}\\
    Beam stiffness law        & $\Kp\!=\!\tfrac{\pi}{4}E r^4/\ell$ & Euler--Bernoulli; branch model of \cite{jacob2024gentle}\\
    Pipe-model exponent $\beta$ & 2.2--2.3 & Area-preserving ($2$) to Murray ($2.5$) \cite{shinozaki1964pipe,mcculloh2003murray,lehnebach2018pipe}\\
    Phyllotactic divergence   & \SI{137.5}{\degree} & Golden angle \cite{okabe2015phyllotaxis,smith2006}\\
    Scaffold crotch angle     & \SIrange{42}{65}{\degree} & Strong-crotch / fruit-bud range (nominal, horticultural practice)\\
    Tree height               & \SI{2.4}{\metre} & \emph{Sim.\ choice} (arm reach); modern dwarf/semi-dwarf $\sim$3--3.5\,m \cite{robinson2006training}\\
    Trunk radius              & \SI{0.035}{\metre} & Mature M.9 trunk $\sim$5--7.5\,cm dia.\ (nominal)\\
    \midrule
    \multicolumn{3}{@{}l}{\emph{Fruit}}\\
    Stem detachment force     & \SIrange{14}{23}{\newton} & Within reported \SIrange{9}{40}{\newton} spread \cite{bu2020detachment,li2016characterizing,param1991design}\\
    Apple mass                & \SI{0.16}{\kilo\gram} ([0.09,0.40]) & Commercial dessert apple $\sim$150--180\,g (nominal grade weight)\\
    Apple diameter            & 4.8--7\,cm & \emph{Sim.\ choice} (Franka 80\,mm gripper); real $\sim$58--92\,mm\\
    Pedicel length            & \SI{25}{\milli\metre} & Pomological descriptions (nominal)\\
    Leaf blade                & $8.5\!\times\!5$\,cm & Elliptic-ovate, L:W$\approx$1.7 (nominal)\\
    \midrule
    \multicolumn{3}{@{}l}{\emph{Platform \& environment}}\\
    Base max speed            & \SI{1.1}{\metre\per\second} & Clearpath Ridgeback datasheet \cite{ridgeback_datasheet}\\
    Base drive effort limit   & \SI{900}{\newton} & Traction $\mu m g$, 135\,kg, $\mu\!\approx\!0.7$ (datasheet \cite{ridgeback_datasheet} + \emph{solver})\\
    Depth camera FOV / range / rate & \SI{75}{\degree} / 0.3--3\,m / \SI{5}{\hertz} & Consumer RGB-D, Intel RealSense D435i class \cite{realsense_d435i}\\
    Terrain roughness         & \SI{3}{\centi\metre} (up to \SI{20}{\centi\metre}) & Agricultural soil random-roughness \cite{verhoest2008roughness}\\
    \bottomrule
  \end{tabular}
\end{table*}

\section{Numerical Stability}
\label{app:stability}
Interactive and applied forces (the mouse pick, the autonomous grip spring, the
twig brush) are made unconditionally stable by three rails on each body's
accumulated force, none of which affects static loading, bending a branch until
it snaps still sees the full force. (1) An \emph{impulse cap}: an applied force
may not accelerate a body past a speed limit in one substep. (2) A \emph{power
fade}: the component of an applied force along a body's own velocity fades out
above a threshold speed, ``a hand cannot keep pushing something flying away.''
(3) A soft \emph{speed brake} above the limit. These matter because the pick
clamp scales with the \emph{whole articulation's} mass, so the picker may legally
apply a large force to a gram-scale twig; without the rails, the instant a
rupture frees such a twig it would take an explosive velocity in one substep and
NaN the articulation. Gram-scale twigs also take no rigid contacts (their extreme
mass ratio is unresolvable); the robot pushes them aside with a bounded penalty
``brush'' force routed through the same rails. Snapped sub-trees additionally
drop out of collision and have their applied forces disabled, so debris cannot be
re-excited.

\section{Reproduction}
\label{app:repro}
All experiments run on a single NVIDIA RTX~2000 Ada laptop GPU
(\SI{8}{\giga\byte}) with \textsc{Newton}~1.3 / Warp~1.14. Each condition fixes
the random seed so the same domain-randomized population is reused, and metrics
are pooled across parallel environments. The experiment runner executes the
matrix one job at a time, writing per-condition JSON and a combined results
table, and an analysis script regenerates every figure and table from those
files. The picking sweeps share a single base seed so the same
domain-randomized population recurs across conditions (a clean paired design);
the full command matrix, exact seeds, and the official evaluation seed set
accompany the code release at \projurl.

\end{document}